\documentclass[runningheads]{llncs}
 
\usepackage[mobile]{eccv}

\makeatletter
\let\ifreview\ifeccv@review
\makeatother

\usepackage[accsupp]{axessibility}  
\usepackage{booktabs}
\usepackage[misc]{ifsym}
\usepackage{graphicx}
\usepackage{multirow}
\usepackage{pifont}
\usepackage{tabularx}
\usepackage{comment}
\usepackage{xr}
\usepackage{xspace} 
\usepackage[table]{xcolor} 

\usepackage[pagebackref,breaklinks,colorlinks,citecolor=eccvblue]{hyperref}

\usepackage{orcidlink}

\newcolumntype{Y}{>{\centering\arraybackslash}X}
\newcommand{\OM}{MonoArt\xspace}
\newcommand{\eg}{\textit{e.g.}}
\newcommand{\ie}{\textit{i.e.}}
\newcommand{\darr}{$\downarrow$\xspace}
\newcommand{\uarr}{$\uparrow$\xspace}
\newcommand{\cmark}{\textcolor{green}{\ding{51}}}
\newcommand{\xmark}{\textcolor{red}{\ding{55}}}

\renewcommand{\paragraph}[1]{%
  \par\medskip
  \noindent\textbf{#1.\ }%
}

\begin{document}

\title{\OM:
Progressive Structural Reasoning for Monocular Articulated 3D Reconstruction}
\titlerunning{\OM: Monocular Articulated 3D Object Reconstruction}

\author{
Haitian Li\textsuperscript{*}~\orcidlink{0009-0003-1229-9206} \and
Haozhe Xie\textsuperscript{*}~\orcidlink{0000-0001-9596-5179} \and
Junxiang Xu \orcidlink{0009-0003-3237-9764} \and\\
Beichen Wen \orcidlink{0009-0008-2890-8332} \and
Fangzhou Hong \orcidlink{0000-0003-2412-1141} \and
Ziwei Liu~\textsuperscript{\Letter}~\orcidlink{0000-0002-4220-5958}}

\authorrunning{H. Li et al.}

\institute{%
S-Lab, Nanyang Technological University, 637335 Singapore\\
\url{https://lihaitian.com/MonoArt}}

\maketitle
\ifreview \else
\def\thefootnote{}
\footnotetext{%
\textsuperscript{*} Equal Contribution
\hspace{4 mm}
\textsuperscript{\Letter} Corresponding Author}
\fi

\begin{figure}
  \ifreview \vspace{-4 mm} \else \vspace{-8 mm} \fi
  \includegraphics[width=\textwidth]{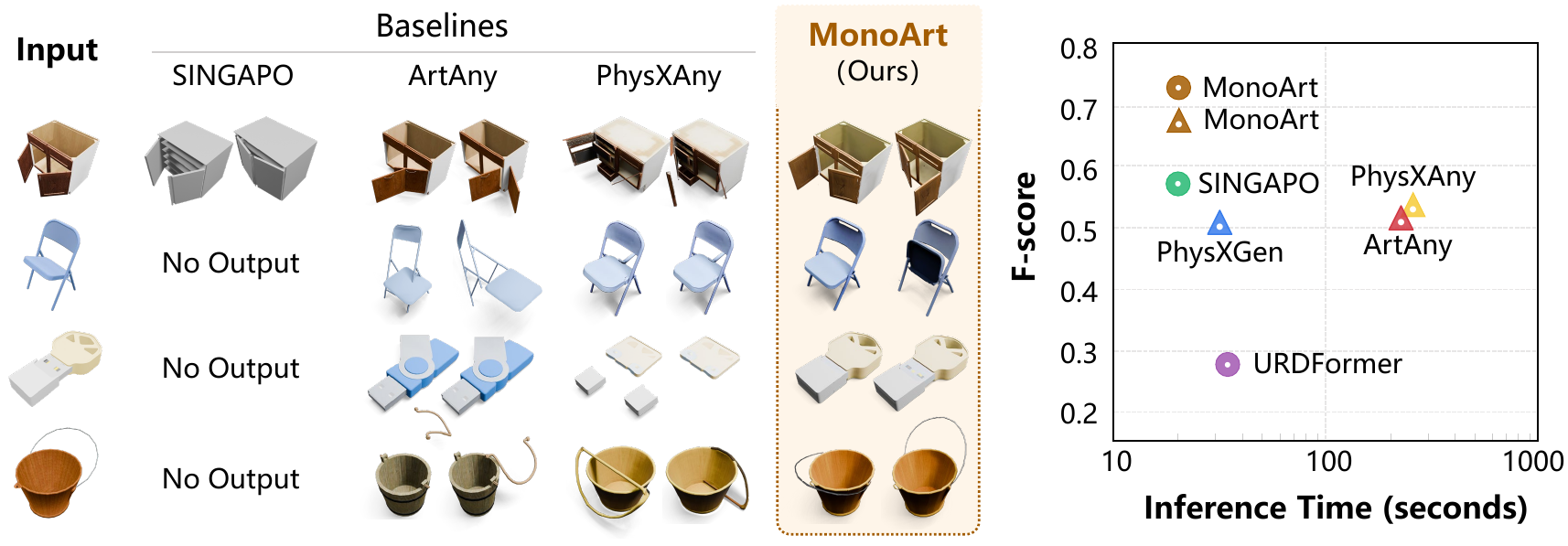}
  \caption{%
  \textbf{(Left)} Qualitative results of SINGAPO~\cite{DBLP:conf/iclr/LiuICSA25}, Articulate-Anything (ArtAny)~\cite{DBLP:conf/iclr/LeXLWYMVKJE25}, PhysX-Anything (PhysXAny)~\cite{DBLP:journals/corr/abs-2511-13648}, and \OM on diverse objects.
  \textbf{(Right)} F-score vs. inference time on the PartNet-Mobility~\cite{DBLP:conf/cvpr/XiangQMXZLLJYWY20} test set. Circles indicate models evaluated on 7 categories, while triangles denote models supporting all 46 categories.}
  \label{fig:teaser}
  \ifreview \vspace{-8 mm} \else \vspace{-12 mm} \fi
\end{figure}

\begin{abstract}

Reconstructing articulated 3D objects from a single image requires jointly inferring object geometry, part structure, and motion parameters from limited visual evidence. 
A key difficulty lies in the entanglement between motion cues and object structure, which makes direct articulation regression unstable.
Existing methods address this challenge through multi-view supervision, retrieval-based assembly, or auxiliary video generation, often sacrificing scalability or efficiency.
We present \OM, a unified framework grounded in progressive structural reasoning. 
Rather than predicting articulation directly from image features, \OM progressively transforms visual observations into canonical geometry, structured part representations, and motion-aware embeddings within a single architecture. 
This structured reasoning process enables stable and interpretable articulation inference without external motion templates or multi-stage pipelines.
Extensive experiments on PartNet-Mobility demonstrate that \OM achieves state-of-the-art performance in both reconstruction accuracy and inference speed. 
The framework further generalizes to robotic manipulation and articulated scene reconstruction.

\end{abstract}

\section{Introduction}

Reconstructing articulated 3D objects from visual observations remains a fundamental challenge in computer vision and graphics. 
Such objects (\eg, laptops, cabinets) are ubiquitous in daily environments, and generating high-fidelity, interactable 3D assets at scale is increasingly vital for applications in robotics~\cite{DBLP:conf/rss/BrohanBCCDFGHHH23, DBLP:conf/rss/BlackBDEEFFG25, DBLP:journals/corr/abs-2601-22153} and scene synthesis~\cite{DBLP:conf/cvpr/YangJZH24, DBLP:conf/cvpr/Xie0H024, DBLP:journals/pami/XieCHL25, DBLP:preprint/arxiv/2505-05474}. 
Despite recent progress in 3D object generation~\cite{DBLP:conf/iclr/TangRZ0Z24, DBLP:conf/cvpr/XiangLXDWZC0Y25, DBLP:conf/cvpr/0009TDCHL0XWSPL25, DBLP:preprint/arxiv/2506-16504}, producing high-quality articulated assets still requires extensive manual effort. 
Unlike rigid reconstruction, articulated reconstruction demands understanding both an object’s structural composition and the kinematic relationships among its parts.

Articulated 3D object modeling has attracted increasing attention in recent years, driving significant advances in reconstruction. 
Existing approaches can be broadly classified into three categories.
Most existing methods~\cite{DBLP:conf/cvpr/WeiCMLZRSNS22, DBLP:conf/iccv/LiuMS23, DBLP:conf/cvpr/WengWTBFGB24, DBLP:conf/iclr/MandiWBS25, DBLP:conf/cvpr/GuoXL0LH25} rely on multiple images captured from videos or frame sequences indicating different articulation states (\eg, open and closed). 
While achieving higher reconstruction accuracy, they require multiple motion states of the same object, which are not always readily available in practice.
To relax this constraint, subsequent approaches~\cite{DBLP:conf/rss/ChenWMMFF024, DBLP:conf/iclr/LiuICSA25, DBLP:conf/iclr/LeXLWYMVKJE25} take a single image as input but reconstruct 3D objects by retrieving and assembling parts from pre-built asset libraries, often leading to texture misalignment and geometric inaccuracies.
Very recent works move beyond retrieval and attempt single-image reconstruction of articulated objects by generating auxiliary videos~\cite{DBLP:conf/siggrapha/LuLTNWWZCH25}, leveraging vision–language models~\cite{DBLP:journals/corr/abs-2511-13648}, or relying on predefined motion directions to infer articulation cues~\cite{DBLP:conf/cvpr/GaoPGZLDT0025, DBLP:conf/iccv/LiZRV25}.
While these approaches demonstrate the potential of single-image articulation modeling, the former is complex and computationally expensive, whereas the latter depends on handcrafted priors that limit generalization.
This reliance on external cues reflects the lack of intrinsic 3D understanding, making it difficult to infer part composition and spatial relationships directly from a single image.

To address these limitations, we propose \textbf{\OM}, an end-to-end framework for monocular articulated 3D reconstruction grounded in progressive structural reasoning.
Instead of directly regressing articulation parameters from image features, \OM progressively constructs structured part representations and lifts them into motion-aware embeddings, enabling stable and interpretable kinematic prediction.
Specifically,
\textbf{TRELLIS-based 3D Generator} first produces a canonical 3D shape from the input image, providing a stable geometric foundation. 
\textbf{Part-aware Semantic Reasoner} then lifts geometry-aligned point features into globally contextualized part-level embeddings through tri-plane aggregation and transformer-based refinement under 3D structural supervision.
These part-aware features are subsequently processed by a \textbf{Dual-Query Motion Decoder}, which decouples spatial motion anchors and semantic part representations, and performs iterative refinement to reason about component-level motion patterns.
Finally, \textbf{Kinematic Estimator} predicts part-level articulation parameters, including the part centroid, mask, joint type, axis, origin, and motion limits, and infers the kinematic tree structure, yielding a coherent articulated 3D reconstruction.
As illustrated in Figure~\ref{fig:teaser}, \OM achieves both higher F-score and substantially lower inference time than existing approaches, demonstrating that explicit structural priors enable not only more accurate but also more efficient articulated reconstruction.

The contributions are summarized as follows:
\begin{itemize}
\item 
We show that embedding 3D structural priors simplifies single-image articulated reconstruction by eliminating reliance on video generation, handcrafted motion templates, or vision–language priors, enabling reliable part decomposition and articulation reasoning.

\item 
We propose \OM, a unified end-to-end framework that progressively reasons from geometry to kinematics, disentangling shape recovery, part-aware encoding, motion decoding, and kinematic regression for stable and physically meaningful articulation inference.

\item
\OM achieves state-of-the-art performance on PartNet-Mobility in both geometric and articulation metrics while significantly reducing inference time. 
The framework further generalizes to robotic manipulation and indoor 3D reconstruction tasks.

\end{itemize}

\section{Related Work}

\paragraph{Articulated Object Modeling}
Articulated object modeling aims to recover object geometry together with part structure and motion from visual observations. 
Early multi-view methods~\cite{DBLP:conf/cvpr/WeiCMLZRSNS22, DBLP:conf/cvpr/SongWFLL24} adopt neural implicit representations to model category-level articulated shapes as deformations of a canonical template, enabling view synthesis but without explicit part or kinematic modeling.
Later works introduce explicit part decomposition and motion reasoning. 
PARIS~\cite{DBLP:conf/iccv/LiuMS23} aligns reconstructions across articulation states to estimate rigid parts and transformations. DTA~\cite{DBLP:conf/cvpr/WengWTBFGB24} and ArticulatedGS~\cite{DBLP:conf/cvpr/GuoXL0LH25} jointly model geometry, segmentation, and articulation parameters from multi-view RGB-D or Gaussian Splatting, producing motion-ready digital twins with high geometric fidelity.
More recent approaches reduce input requirements by leveraging generative or language priors. 
FreeArt3D~\cite{DBLP:conf/siggrapha/0003LW0L25} optimizes geometry from sparse articulated views, while SINGAPO~\cite{DBLP:conf/iclr/LiuICSA25}, NAP~\cite{DBLP:conf/nips/LeiDSGD23}, and MeshArt~\cite{DBLP:conf/cvpr/GaoSLD25} predict articulation trees and synthesize articulated parts. 
Articulate-Anything~\cite{DBLP:conf/iclr/LeXLWYMVKJE25} formulates the task as vision–language reasoning to infer symbolic part hierarchies, and PhysX-Anything~\cite{DBLP:journals/corr/abs-2511-13648} further incorporates VLM priors to predict physically plausible structures and interactions. 
These methods improve scalability and controllability, often at the expense of precise instance-level reconstruction.

\paragraph{3D Part Segmentation}
3D part segmentation aims to decompose objects into semantic parts. 
Early learning-based methods~\cite{DBLP:conf/nips/LiBSWDC18, DBLP:conf/cvpr/QiSMG17, DBLP:conf/nips/QiYSG17, DBLP:conf/nips/QianLPMHEG22, DBLP:conf/iccv/ZhaoJJTK21, DBLP:journals/tog/HanockaHFGFC19, DBLP:conf/eccv/LiYZ22, DBLP:journals/ijcv/LiuXZYJNT25} focus on fully supervised point- or mesh-level classification using curated 3D datasets~\cite{DBLP:journals/tog/ChenGF09, DBLP:journals/corr/ChangFGHHLSSSSX15, DBLP:journals/tog/YiKCSYSLHSG16, DBLP:conf/cvpr/MoZCYTGS19}. 
While effective, these approaches are limited by the scale and diversity of part annotations.
To improve open-world generalization and enable zero-shot capabilities, recent methods leverage 2D foundation vision models~\cite{DBLP:conf/iccv/KirillovMRMRGXW23, DBLP:conf/cvpr/LiZZYLZWYZHCG22, DBLP:journals/tmlr/OquabDMVSKFHMEA24, DBLP:conf/icml/RadfordKHRGASAM21, DBLP:conf/aaai/LiuSXLLZ25}. 
PartSLIP~\cite{DBLP:conf/cvpr/LiuZCHLP023}, PartSLIP++~\cite{DBLP:journals/corr/abs-2312-03015}, and ZeroPS~\cite{DBLP:conf/3dim/XueCLS25} transfer image–language and segmentation priors to 3D via multi-view reasoning or prompt-based inference, while PartDistill~\cite{DBLP:conf/cvpr/UmamYCCL24}, SaMesh~\cite{DBLP:journals/corr/abs-2408-13679}, and SAMPart3D~\cite{DBLP:journals/corr/abs-2411-07184} distill 2D foundation features into geometric representations.
More recently, scaling-based approaches train feed-forward 3D segmentation models with large-scale part annotations. 
Find3D~\cite{DBLP:conf/iccv/MaYG25} leverages foundation models for pseudo-label generation, PartField~\cite{DBLP:conf/iccv/LiuUXSFSG25} learns ambiguity-aware continuous feature fields, and P3-SAM~\cite{DBLP:journals/corr/abs-2509-06784} and PartSAM~\cite{DBLP:conf/iclr/ZhuWXZCDLLW26} demonstrate that large-scale part supervision yields strong point-wise representations for part prompting.

\section{Our Approach}

\subsection{Overview}

\begin{figure}[!t]
  \includegraphics[width=1.0\textwidth]{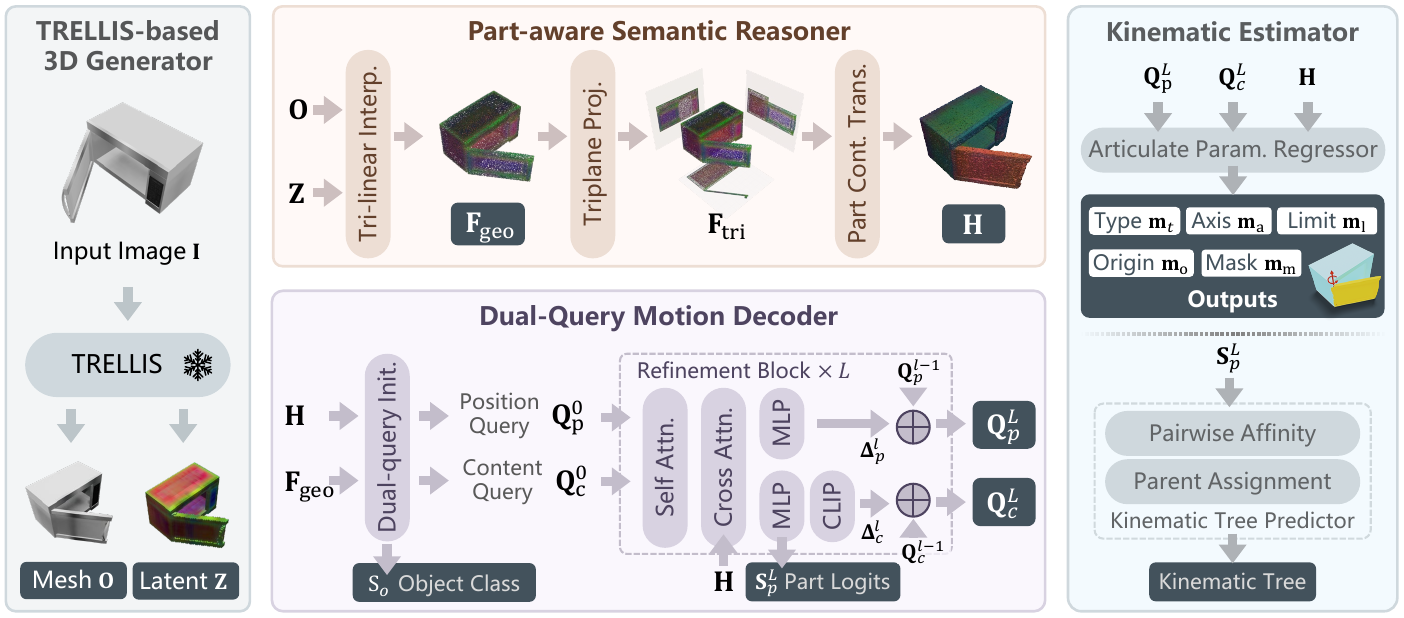}
  \caption{\textbf{Overview of \OM.} 
  TRELLIS-based 3D Generator reconstructs a canonical shape from a single image. 
  Part-Aware Semantic Reasoner derives tri-plane-based part embeddings. 
  Dual-Query Motion Decoder performs iterative motion reasoning, and 
  Kinematic Estimator predicts part-level articulation parameters (motion type, origin, axis, limits) and infers the kinematic tree structure.
  Note that ``Attn.'', ``Interp.'', ``Proj.'', ``Cont.'', ``Trans.'', and ``Init.'' represent ``Attention'', ``Interpolation'', ``Projection'', ``Contrast'', ``Transformer'', and ``Initialization'', respectively.
  $\oplus$ and $\otimes$ denote element-wise addition and matrix multiplication, respectively.}
  \label{fig:method}
  \vspace{-6 mm}
\end{figure}

\OM reconstructs articulated 3D objects from a single image by progressively transforming visual observations into geometry-aware, part-aware, and motion-aware representations.
As shown in Fig.~\ref{fig:method}, the framework consists of four components: 
TRELLIS-based 3D Generator (Sec.~\ref{sec:3D-generator}), 
Part-Aware Semantic Reasoner (Sec.~\ref{sec:pasr}), 
Dual-Query Motion Decoder (Sec.~\ref{sec:dqmd}), and 
Kinematic Estimator (Sec.~\ref{sec:ke}).

Given an input image $\mathbf{I}$, TRELLIS-based 3D Generator reconstructs a canonical geometry $\mathbf{O}$ using a frozen TRELLIS backbone and produces geometry-aligned latent features $\mathbf{Z}$.
Based on $(\mathbf{O}, \mathbf{Z})$, Part-Aware Semantic Reasoner derives part-aware features $\mathbf{H}$ that encode explicit part decomposition guided by 3D structural annotations.
Then, Dual-Query Motion Decoder initializes position and content queries $(\mathbf{Q}_p^{0}, \mathbf{Q}_c^{0})$ from $(\mathbf{H}, \mathbf{F}_\text{geo})$ and iteratively refines them to obtain motion-aware representations $(\mathbf{Q}_p^{L}, \mathbf{Q}_c^{L})$, jointly reasoning about part localization and motion semantics.
Finally, Kinematic Estimator transforms the refined queries into explicit articulation parameters, namely part masks $\mathbf{m}_m$, motion types $\mathbf{m}_t$, motion origins $\mathbf{m}_o$, motion axes $\mathbf{m}_a$, and motion limits $\mathbf{m}_l$, while inferring the kinematic hierarchy to produce an articulated 3D representation.

\subsection{TRELLIS-based 3D Generator}
\label{sec:3D-generator}

Given a single RGB image $\mathbf{I}$, \OM first reconstructs a canonical 3D object geometry using TRELLIS~\cite{DBLP:conf/cvpr/XiangLXDWZC0Y25} as a frozen 3D generation backbone. 
TRELLIS predicts a structured sparse voxel latent representation $\mathbf{Z} \in \mathbb{R}^{N_z \times N_z \times N_z \times d_1}$, 
where $N_z$ denotes the voxel grid resolution and each active voxel stores a $d_1$-dimensional feature while empty regions remain inactive, forming a spatially structured sparse volume.
The latent $\mathbf{Z}$ is subsequently decoded by the mesh decoder of TRELLIS to produce an explicit 3D mesh $\mathbf{O}$, which serves as the canonical geometry for downstream part reasoning and articulation inference.




\subsection{Part-Aware Semantic Reasoner}
\label{sec:pasr}

Our goal is to derive part-aware point features $\mathbf{H}$ from the canonical geometry $\mathbf{O}$ and the sparse voxel latent $\mathbf{Z}$ predicted by TRELLIS-based 3D generator, so that downstream modules can reason about object parts and their motions.

\paragraph{Tri-linear Interpolation}
We sample $M$ point sets $\{\mathbf{p}_m\}_{m=1}^{M}$ on the surface of $\mathbf{O}$, where $\mathbf{p}_m \in \mathbb{R}^3$.
Given the sparse voxel feature volume $\mathbf{Z}$, we obtain point-aligned features by tri-linear interpolation over the neighboring voxels:
\begin{equation}
  \mathbf{f}_m = \mathrm{TrilinearInterp}(\mathbf{Z}, \mathbf{p}_m),
  \qquad
  \mathbf{F}_{\text{geo}} = \{\mathbf{f}_m\}_{m=1}^{M}.
\end{equation}
Here, each point $\mathbf{p}_m$ is first mapped to its corresponding continuous voxel coordinate in the $N_z \times N_z \times N_z$ grid.
The feature $\mathbf{f}_m \in \mathbb{R}^{d_1}$ is then computed as the weighted combination of the eight neighboring voxel features according to standard tri-linear interpolation~\cite{DBLP:conf/eccv/XieYZMZS20}.
This converts the discrete voxel latent $\mathbf{Z}$ into continuous surface-aligned features $\mathbf{F}_{\text{geo}} \in \mathbb{R}^{M \times d_1}$.

\paragraph{Triplane Projection}
To incorporate global spatial context while preserving geometric structure, we project the geometry-aligned point features $\mathbf{F}_{\text{geo}}$ onto three orthogonal planes (XY, YZ, ZX) following the triplane formulation~\cite{DBLP:conf/cvpr/ChanLCNPMGGTKKW22}.
Specifically, each surface point $\mathbf{p}_m$ is orthographically projected onto the three planes according to its 3D coordinates, and its feature $\mathbf{f}_m$ is accumulated to the corresponding 2D grid locations via bilinear interpolation. 
This yields three feature maps of resolution $N_t \times N_t$, forming $\mathbf{F}_{\text{tri}} \in \mathbb{R}^{3 \times N_t \times N_t \times d_1}$.
%

\paragraph{Part Contrast Transformer}
The triplane features $\mathbf{F}_{\text{tri}}$ are flattened into a token sequence $\mathbf{T} \in \mathbb{R}^{3N_t^2 \times d_1}$ and processed by a multi-layer self-attention Transformer to capture global interactions across planes, producing refined tokens $\mathbf{T}'$.
The refined tokens are reshaped back into triplane feature maps $\mathbf{F}'_{\text{tri}} \in \mathbb{R}^{3 \times N_t \times N_t \times d_1}$.
For each surface point $\mathbf{p}_m$, we project its 3D coordinate onto the three orthogonal planes and query the refined triplane feature maps via a tri-plane sampling operation.
This yields the updated point embedding
\begin{equation}
  \mathbf{h}_m = \mathrm{MLP}(\mathrm{TriQuery}(\mathbf{F}'_{\text{tri}}, \mathbf{p}_m)),
  \qquad
  \mathbf{H} = \{\mathbf{h}_m\}_{m=1}^{M},
\end{equation}
where 
TriQuery projects $\mathbf{p}_m$ onto the XY/YZ/ZX planes, bilinearly samples each plane feature map, and concatenates the three sampled features.
Then, the MLP then lifts the aggregated feature from dimension $d_1$ to $d_2$.
These embeddings encode part-aware structural representations and are supervised by 3D part annotations using triplet loss~\cite{DBLP:conf/cvpr/SchroffKP15}.

\subsection{Dual-Query Motion Decoder}
\label{sec:dqmd}

Articulated reasoning requires jointly modeling what constitutes a movable part and where its motion is spatially anchored.
To this end, we adopt a dual-query formulation that disentangles semantic representation and geometric localization via two complementary query types: a content query $\mathbf{Q}_c \in \mathbb{R}^{N_q \times d_2}$ encoding part semantics, and a position query $\mathbf{Q}_p \in \mathbb{R}^{N_q \times 3}$ representing spatial motion anchors,
where $N_q$ denotes the number of dual queries.
The dual queries are initialized from global object context and subsequently refined through $L$ stacked refinement blocks in an iterative manner.

\paragraph{Dual-query Initialization}
Given the part-aware point embeddings $\mathbf{H}$ and geometry-aligned features $\mathbf{F}_\text{geo}$, we first aggregate global object context by applying global pooling over $\mathbf{H}$ and $\mathbf{F}_\text{geo}$, followed by feature concatenation to obtain an object-level representation.
This aggregated feature is projected to generate the initial content queries $\mathbf{Q}_c^0 \in \mathbb{R}^{N_q \times d_2}$ and position queries $\mathbf{Q}_p^0 \in \mathbb{R}^{N_q \times 3}$, together with an auxiliary object-category prediction ${S}_o$. 

\paragraph{Refinement Block}
The refinement block iteratively updates the dual queries over $L$ layers to progressively refine articulation hypotheses.
Given the initialized position query $\mathbf{Q}_p^0 \in \mathbb{R}^{N_q \times 3}$ and content query $\mathbf{Q}_c^0$, each layer first applies self-attention to model inter-part interactions, followed by cross-attention where the queries attend to the visual feature $\mathbf{H}$ to retrieve image evidence.
For the $l$-th layer, the position query $\mathbf{Q}_p^l$ represents spatial motion anchors that localize candidate articulation points.
It is updated via a residual scheme $\mathbf{Q}_p^l = \mathbf{Q}_p^{l-1} + \Delta_p^l$, 
allowing progressive refinement of spatial motion anchors.
In parallel, the content query is also updated in a residual manner $\mathbf{Q}_c^l = \mathbf{Q}_c^{l-1} + \Delta_c^l$,
where $\Delta_c^l$ captures semantic refinement derived from self- and cross-attention.
Based on the refined content queries, we predict per-query part class logits $\mathbf{S}_p^l$, which provide structural supervision and are further used to retrieve semantic prototypes from frozen CLIP text embeddings. 
The retrieved prototypes are fused into the content branch to enhance part-level semantic consistency.

\paragraph{Query Confidence Estimation}
Since the number of dual queries $N_q$ defines an upper bound on articulated components, some queries may correspond to invalid part hypotheses.
To address this, we predict a confidence score $c_i \in [0,1]$ from the refined content embedding $\mathbf{Q}_{c,i}^{L}$ of the $i$-th query, indicating the reliability of its part hypothesis.
During training, confidence supervision is derived from Hungarian matching between predicted part masks and ground-truth components. Matched queries are assigned confidence targets proportional to their mask overlap, while unmatched queries are treated as null hypotheses with zero confidence.
At inference time, queries with confidence below a threshold are discarded, allowing the model to automatically determine the number of parts.

\subsection{Kinematic Estimator}
\label{sec:ke}

Given $\mathbf{Q}_p^L$, $\mathbf{Q}_c^L$, part logits $\mathbf{S}_p^L$, and point embedding $\mathbf{H}$, Kinematic Estimator predicts articulation parameters and a kinematic tree. 

\paragraph{Articulate Parameter Regressor}
The refined dual queries are $\mathbf{Q}_p^L \in \mathbb{R}^{N_q \times 3}$ and $\mathbf{Q}_c^L \in \mathbb{R}^{N_q \times d_2}$,
where $\mathbf{Q}_p^L$ is interpreted as part centroid, \ie, the spatial center of an articulated component.

\noindent \textbf{Part mask} is obtained by query–point matching.
Specifically, we compute the affinity between content queries and point features $\mathbf{H} \times \mathbb{R}^{M \times d_2}$:
\begin{equation}
  \mathbf{m}_m = \mathbf{Q}_c^L \mathbf{H}^\top,
  \qquad
  \mathbf{m}_m \in \mathbb{R}^{N_q \times M}.
\end{equation}
Each row of $\mathbf{m}_m$ indicates the soft assignment of surface points to a query-induced articulated part.

\noindent \textbf{Joint parameters} are regressed from query-level representations.
We integrate $\mathbf{Q}_p^L$, $\mathbf{Q}_c^L$, and part-level features $\mathbf{H}$ into a unified per-query representation, which is processed by lightweight MLP heads to predict physically interpretable joint parameters:
\begin{itemize}
\item \textbf{Joint type} $\mathbf{m}_t \in \mathbb{R}^{N_q \times N_t}$, where $N_t$ denotes the number of predefined motion categories (\eg, fixed, revolute, prismatic, and continuous);

\item \textbf{Joint axis} $\mathbf{m}_a \in \mathbb{R}^{N_q \times 3}$, representing the unit direction vector of the predicted joint axis;

\item \textbf{Joint pivot} $\mathbf{m}_o \in \mathbb{R}^{N_q \times 3}$, denoting the pivot point through which the motion axis passes;

\item \textbf{Joint limits} $\mathbf{m}_l \in \mathbb{R}^{N_q \times 2}$, parameterized by a center and a symmetric span.
\end{itemize}

\paragraph{Kinematic Tree Predictor}
The articulated objects are modeled as tree-structured kinematic graphs.
We use $i \in \{1, ..., N_q\}$ to index queries.
The part logits are given by $\mathbf{S}_p^{L} \in \mathbb{R}^{N_q \times N_c}$, and 
the predicted category distribution of part $i$ is denoted as $\mathbf{s}_i \in \mathbb{R}^{N_c}$.

\noindent \textbf{Pairwise affinity} is computed to model potential parent–child relations among predicted parts.
For each ordered pair $(i,j)$, we compute a semantic attachment score using a learnable compatibility matrix $\mathbf{C} \in \mathbb{R}^{N_c \times N_c}$:
\begin{equation}
  \mathbf{S}_{i, j} = \mathbf{s}_i^\top\mathbf{C}\mathbf{s}_j.
\end{equation}
This bilinear formulation captures category-level attachment priors in a data-driven manner.
To obtain the probability that part $j$ serves as the parent of part $i$, we normalize the scores over all candidate parents:
\begin{equation}
    P(j | i) = \mathrm{Softmax}(\mathbf{S}_{i, j}).
\end{equation}

\noindent \textbf{Parent assignment} selects, for each part $i$, the parent with the highest attachment probability $P(j | i)$, optionally including a learnable root node as a candidate parent. 
To ensure a valid kinematic hierarchy, we enforce a single-root, cycle-free constraint during structure construction.

\section{Experiments}

\subsection{Evaluation Protocol}

\paragraph{Dataset}
We use PartNet-Mobility~\cite{DBLP:conf/cvpr/XiangQMXZLLJYWY20} as the benchmark, which contains approximately 2K articulated objects with part-level geometry and joint annotations, covering fixed, prismatic, revolute, and continuous joints.
We adopt two evaluation splits: 
\textbf{(1)} a 7-category setting following SINGAPO~\cite{DBLP:conf/iclr/LiuICSA25} (Storage, Table, Refrigerator, Dishwasher, Oven, Washer, and Microwave), and 
\textbf{(2)} a full 46-category setting following PhysX-Anything~\cite{DBLP:journals/corr/abs-2511-13648} to evaluate large-scale multi-class generalization.

\paragraph{Metrics}
Following FreeArt3D~\cite{DBLP:conf/siggrapha/0003LW0L25},
we evaluate both motion-aware geometric reconstruction quality
and kinematic prediction accuracy.

\noindent \textbf{Geometric reconstruction quality} includes four metrics: Chamfer Distance (CD), F-Score, PSNR, and CLIP similarity.
For each shape, we uniformly sample six articulation states along the predicted motion range and generate the corresponding meshes using the estimated joint parameters.
All metrics are computed at each state and averaged across states. 
Predicted meshes are aligned with ground-truth meshes using~\cite{DBLP:conf/cvpr/LiuSCZXW00G024} prior to evaluation.
CD and F-Score (threshold 0.05) are computed on 100k uniformly sampled surface points per aligned mesh.
For appearance, we render both predicted and ground-truth meshes at the six articulation states, sampling 10 random viewpoints per state from a unit sphere (60 images in total), and compute the average PSNR and CLIP similarity~\cite{DBLP:conf/icml/RadfordKHRGASAM21} using ViT-L/14@336px.

\noindent \textbf{Kinematic prediction accuracy} includes three metrics: Type Accuracy, Axis Direction Error, and Pivot Distance Error.
Predicted parts are first matched to ground-truth parts with bipartite matching to establish one-to-one correspondence.
Type Accuracy measures joint type classification correctness.
Axis Direction Error $e_\text{axis}$ is defined as the angular deviation between predicted and ground-truth motion axes $\mathbf{a}_p$ and $\mathbf{a}_g$ for both revolute and prismatic joints:
\begin{equation}
e_\text{axis} =
\min \left(
\arccos
\left(
\frac{\mathbf{a}_p \cdot \mathbf{a}_g}
{\lVert\mathbf{a}_p\rVert_2\lVert\mathbf{a}_g\rVert_2}
\right),
\arccos
\left(
\frac{-\mathbf{a}_p \cdot \mathbf{a}_g}
{\lVert\mathbf{a}_p\rVert_2\lVert\mathbf{a}_g\rVert_2}
\right)
\right).
\end{equation}
Pivot Distance Error $e_{\text{pivot}}$ measures the distance between the predicted joint pivot $\mathbf{o}_p$ to the ground-truth pivot $\mathbf{o}_g$:
\begin{equation}
e_\text{pivot} = \frac
{|(\mathbf{o}_p - \mathbf{o}_g)\cdot(\mathbf{a}_p \times \mathbf{a}_g)|}
{|\mathbf{a}_p \times \mathbf{a}_g|}.
\end{equation}
All geometric quantities are evaluated in normalized object coordinates.

\begin{table}[!t]
  \caption{%
  \textbf{Quantitative comparison on the PartNet-Mobility dataset.} 
  CD is scaled by $\times 10^{-2}$.
  Type Acc. (\%) denotes joint type classification accuracy.
  Axis Err. (rad) represents joint axis direction error..
  Pivot Err. is joint pivot distance error in normalized object coordinates.
  Best results are highlighted in bold.}
  \label{tab:main_results}
  \vspace{-2 mm}
  \begin{tabularx}{\linewidth}{lccccccc}
    \toprule
    \multirow{2}{*}{Method} & 
    \multicolumn{4}{c}{Geometry} &
    \multicolumn{3}{c}{Kinematics} \\
    \cmidrule(lr){2-5} \cmidrule(lr){6-8} 
    & 
    CD \darr & 
    F-Score \uarr & 
    PSNR \uarr & 
    CLIP \uarr &
    Type Acc. \uarr & 
    Axis Err. \darr & 
    Pivot Err. \darr \\
    \midrule
    \rowcolor{gray!10} 
    \multicolumn{8}{l}{\textit{Partial 7 classes}} \\
    URDFormer~\cite{DBLP:conf/rss/ChenWMMFF024} & 
    4.73       & 0.275      & 12.43      & 0.845 &
    35.22      & 1.324     & 0.404 \\
    SINGAPO~\cite{DBLP:conf/iclr/LiuICSA25} & 
    1.26       & 0.572      & 15.22      & 0.870 &
    77.12      & 0.493      & 0.201 \\
    \bf{\OM} & 
    \bf{0.77}  & \bf{0.728} & \bf{17.55} & \bf{0.926} &
    \bf{88.26} & \bf{0.209} & \bf{0.085} \\
    \midrule
    \rowcolor{gray!10} 
    \multicolumn{8}{l}{\textit{All 46 classes}} \\
    ArtAny~\cite{DBLP:conf/iclr/LeXLWYMVKJE25} & 
    2.07       & 0.514      & 16.44      & 0.866 &
    43.32      & 0.440      & 0.347 \\
    PhysXGen~\cite{DBLP:journals/corr/abs-2507-12465} & 
    3.06      & 0.501      & 16.38      & 0.859 &
    46.82      & 0.941      & 0.208  \\
    PhysXAny~\cite{DBLP:journals/corr/abs-2511-13648} & 
    1.88       & 0.531      & 17.07      & 0.880 &
    63.35      & \bf{0.289} & 0.173 \\
    \bf{\OM} & 
    \bf{1.25}  & \bf{0.670} & \bf{18.55} & \bf{0.907} &
    \bf{67.47} & 0.423      & \bf{0.108} \\
    \bottomrule
  \end{tabularx}
  \vspace{-4 mm}
\end{table}

\begin{figure}[!t]
  \includegraphics[width=\textwidth]{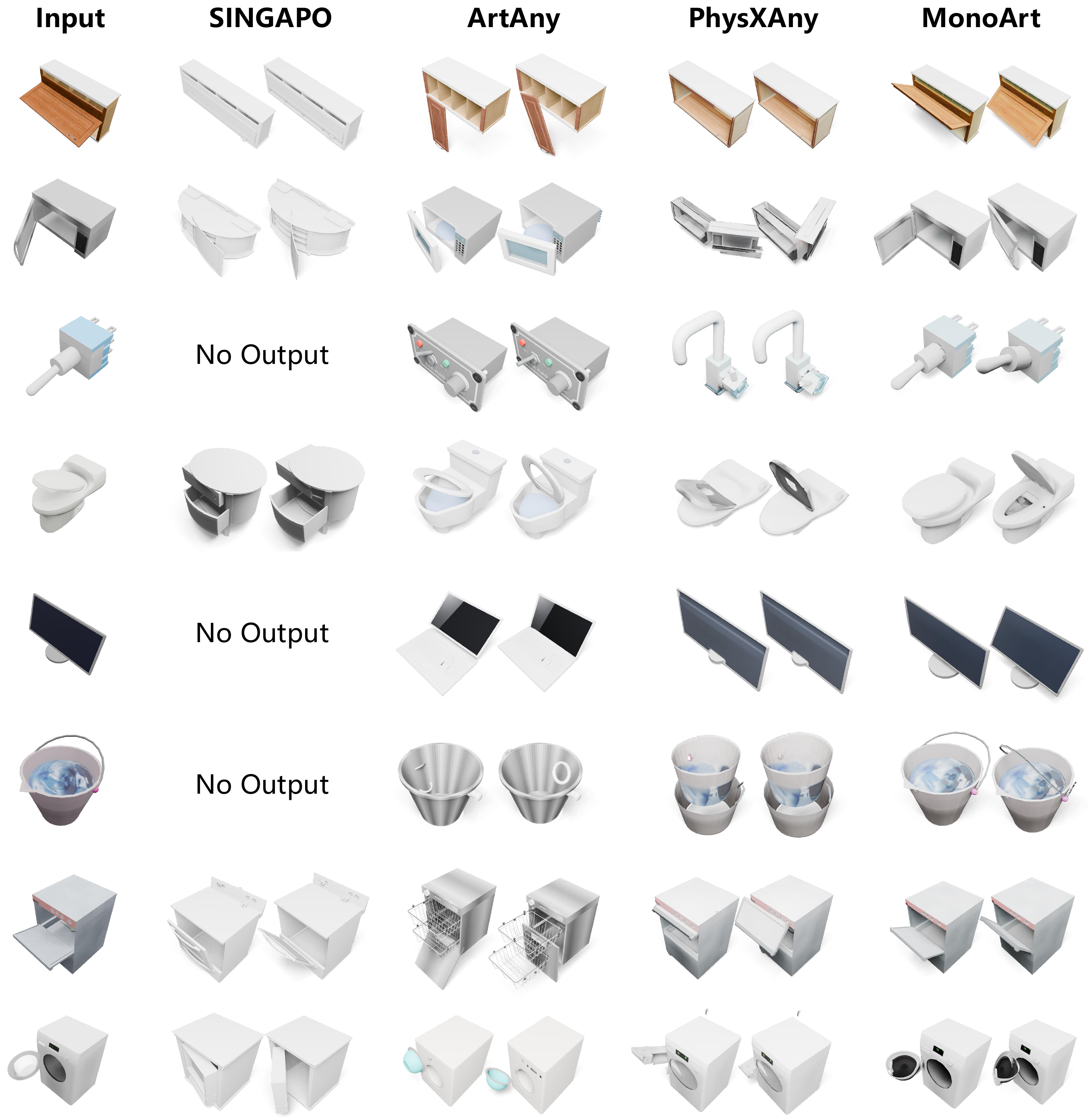}
  \caption{\textbf{Qualitative results on the test set of PartNet-Mobility.} ArtAny and PhysXAny denote Articulate-Anything and PhysXAnything, respectively. For each object, we show the reconstructed geometry under two sampled articulated states.}
  \label{fig:partnet}
  \vspace{-4 mm}
\end{figure}

\subsection{Implementation Details}

\paragraph{Hyperparameters}
We sample $M = 100{,}000$ surface points from the reconstructed mesh for part-aware reasoning. 
The structured voxel latent resolution is $N_z = 64$. 
The tri-plane resolution is $N_t = 128$ with feature dimension $d_1 = 8$, and the lifted point embedding dimension is $d_2 = 448$. 
We use $N_q = 100$ dual queries and a 6-layer dual-query motion decoder $(L = 6)$.

\paragraph{Training Procedure}
We adopt a four-phase training strategy:
\textbf{1)} Warm up Part-Aware Semantic Reasoner with triplet supervision to learn motion-aware part embeddings.
\textbf{2)} Freeze Part-Aware Semantic Reasoner and train the dual-query initialization branch using object-category supervision.
\textbf{3)} Jointly optimize Part-Aware Semantic Reasoner, Dual-query Motion Decoder, and Articulation Parameter Regressor.
\textbf{4)} Freeze all preceding modules and train Kinematic Tree Predictor to model parent–child relations.
Additional training hyperparameters and loss configurations are provided in the Appendix.

\subsection{Main Results}

We compare \OM with state-of-the-art monocular articulated object reconstruction methods, including
URDFormer~\cite{DBLP:conf/rss/ChenWMMFF024}, SINGAPO~\cite{DBLP:conf/iclr/LiuICSA25}, Articulate-Anything~\cite{DBLP:conf/iclr/LeXLWYMVKJE25}, PhysXGen~\cite{DBLP:journals/corr/abs-2507-12465}, and PhysXAnything~\cite{DBLP:journals/corr/abs-2511-13648}.
For retrieval-based baselines (Articulate-Anything and SINGAPO), we remove ground-truth test shapes from their retrieval database to ensure fair comparison and avoid data leakage.

\paragraph{PartNet-Mobility Benchmark}
On PartNet-Mobility~\cite{DBLP:conf/cvpr/XiangQMXZLLJYWY20}, \OM achieves the best overall performance in both geometry reconstruction and kinematic prediction.
For the partial 7 classes, it substantially outperforms prior methods with clear margins in reconstruction quality and articulation estimation, particularly showing large gains in joint type accuracy and significant reductions in axis and pivot errors.
On the full 46 classes, \OM remains consistently superior, delivering the strongest overall reconstruction fidelity while achieving the highest joint type accuracy and the lowest pivot error. 
Notably, it reduces pivot error by more than 40\%, demonstrating robust structural reasoning across diverse articulated categories.
Qualitative results in Fig.~\ref{fig:partnet} further show that \OM produces more faithful geometry and more accurate joint predictions, leading to more plausible articulated motion.

\begin{figure}[!t]
  \includegraphics[width=\textwidth]{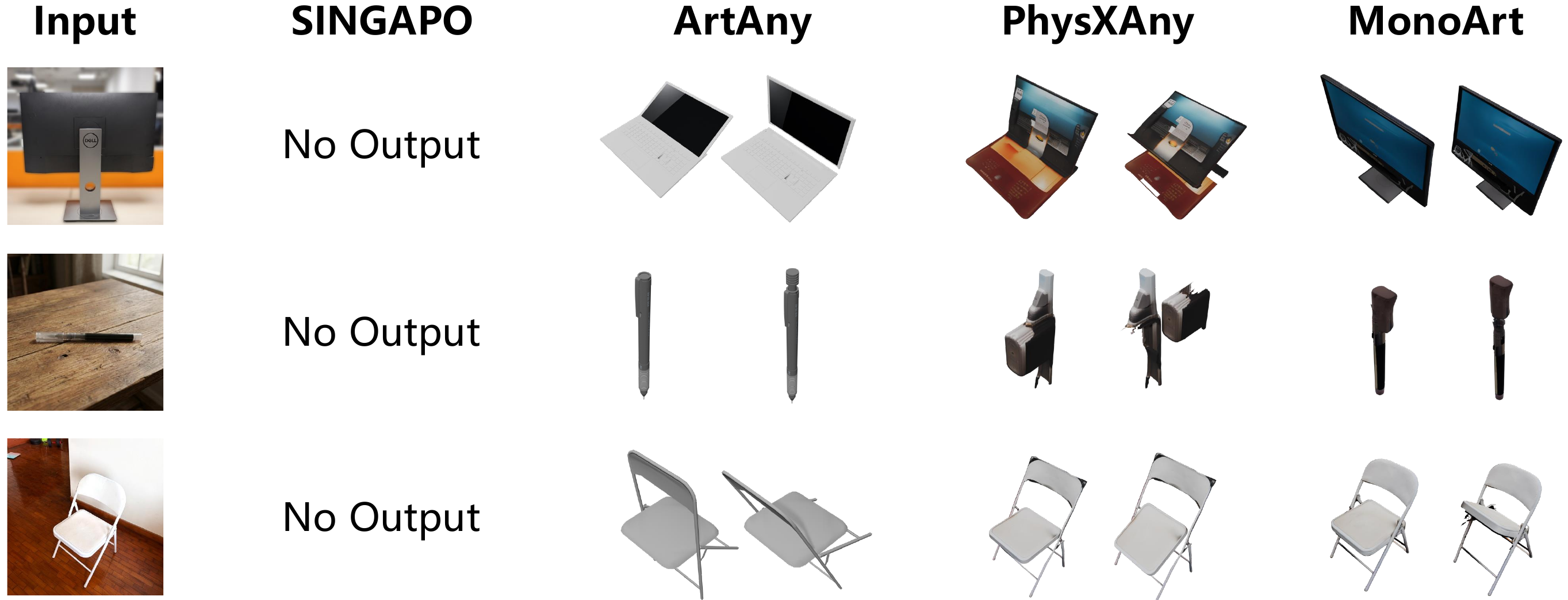}
  \caption{\textbf{Qualitative results on in-the-wild images.} ArtAny and PhysXAny denote Articulate-Anything and PhysXAnything, respectively. For each object, we show the reconstructed geometry under two sampled articulated states.}
  \label{fig:real-world}
  \vspace{-4 mm}
\end{figure}

\paragraph{Real-World Generalization}
To evaluate real-world generalization, we collect approximately 100 in-the-wild images from the Internet using category keywords, covering common everyday articulated objects.
As shown in Fig.~\ref{fig:real-world}, \OM produces coherent geometry and plausible articulation across diverse real-world appearances, despite being trained primarily on synthetic data.
We conduct a user study with 20 participants to evaluate geometric and kinematic quality of the generated articulations. 
Participants rate rendered videos on two 1–5 scales. 
Our method achieves the highest scores (4.63 / 4.37), outperforming PhysX-Anything (3.34 / 3.12), SINGAPO (2.55 / 2.87), Articulate-Anything (2.72 / 2.60), PhysXGen (2.53 / 2.46), and URDFormer (1.37 / 1.49).

\subsection{Ablation Study}

We perform ablations on \OM to study key design choices in Part-Aware Semantic Reasoner, Dual-Query Motion Decoder, and Kinematic Estimator. 
All experiments are conducted on PartNet-Mobility (46 classes), evaluating their impact on geometry and kinematics.

\paragraph{Part-Aware Semantic Reasoner}
As shown in Table~\ref{tab:ablation-pasr}, we ablate the design of the Part-Aware Semantic Reasoner.
Removing the reasoner significantly degrades both geometry and kinematics, especially joint type accuracy and pivot prediction, confirming its critical role in motion-aware part reasoning. 
In this variant, voxel features and lifted point embeddings are directly fused via trilinear interpolation followed by an MLP.
To make the learned feature $\mathbf{H}$ part-aware, we supervise the reasoner with a triplet loss that enforces part-level feature separation. 
Replacing it with cross-entropy supervision leads to clear performance drops, and removing the loss further degrades results. 
Triplet supervision consistently achieves the best performance across all metrics, highlighting the importance for discriminative and motion-consistent part representations.
%
%

\begin{table}[!t]
  \caption{\textbf{Ablation of Part-Aware Semantic Reasoner.} 
  CD is scaled by $\times 10^{-2}$. 
  Type Acc. (\%) is joint classification accuracy. 
  Axis Err. (rad) and Pivot Err. denote axis direction and pivot distance errors.
  Note that ``CE'' and ``Tri.'' denote ``Cross-Entropy'' and ``Triplet'', respectively.}
  \label{tab:ablation-pasr}
  \vspace{-2 mm}
  \begin{tabularx}{\linewidth}{YcYYYccc}
    \toprule
    \multirow{2}{*}{Enabled} &
    \multirow{2}{*}{Loss} & 
    \multicolumn{3}{c}{Geometry} &
    \multicolumn{3}{c}{Kinematics} \\
    \cmidrule(lr){3-5} \cmidrule(lr){6-8} 
    & & 
    CD \darr & 
    F-Score \uarr & 
    PSNR \uarr & 
    Type Acc. \uarr & 
    Axis Err. \darr & 
    Pivot Err. \darr \\
    \midrule
    \xmark & \xmark & 
    1.74       & 0.626      & 17.96 & 
    24.72      & 0.549      & 0.237 \\
    \cmark & \xmark & 
    1.63       & 0.643      & 17.74 & 
    41.60      & 0.922      & 0.323 \\
    \cmark & CE & 
    1.49       & 0.648      & 17.71 & 
    57.74      & 1.029      & 0.302 \\
    \midrule
    \cmark & Tri. & 
    \bf{1.25}  & \bf{0.670} & \bf{18.55} & 
    \bf{67.47} & \bf{0.423} & \bf{0.108} \\
    \bottomrule
  \end{tabularx}
\end{table}

\begin{table}[!t]
  \caption{\textbf{Ablation of Dual-Query Motion Decoder.} 
  CD is scaled by $\times 10^{-2}$. 
  Type Acc. (\%) is joint classification accuracy. 
  Axis Err. (rad) and Pivot Err. denote axis direction and pivot distance errors.
  Note that ``DQI.'' indicates whether Dual-Query Initialization is enabled.
  ``Res.'' denotes residual updates in the position and/or content query branches.
  $L$ is the number of refinement blocks.}
  \label{tab:ablation-qdmd}
  \vspace{-2 mm}
  \begin{tabularx}{\linewidth}{YYYYccccc}
    \toprule
    \multirow{2}{*}{DQI.} &
    \multirow{2}{*}{Res.} & 
    \multirow{2}{*}{$L$} & 
    \multicolumn{3}{c}{Geometry} &
    \multicolumn{3}{c}{Kinematics} \\
    \cmidrule(lr){4-6} \cmidrule(lr){7-9} 
    & & &
    CD \darr & 
    F-Score \uarr & 
    PSNR \uarr & 
    Type Acc. \uarr & 
    Axis Err. \darr & 
    Pivot Err. \darr \\
    \midrule
    \xmark & Both & 6 & 
    1.67       & 0.622      & 18.11 & 
    44.06      & 0.472      & 0.329 \\
    \cmark & $Q_p^l$ & 6 & 
    1.73          & 0.640          & 17.81 & 
    66.41          & 0.523          & 0.181 \\
    \cmark & $Q_c^l$ & 6 & 
    1.29       & 0.663      & 17.80 & 
    60.88      & 0.506      & 0.184 \\
    \midrule
    \cmark & Both & 0 & 
    1.70          & 0.652          & 17.94 & 
    62.65          & 0.640          & 0.186 \\
    \cmark & Both & 1 & 
    1.71          & 0.652          & 17.91 & 
    63.12          & 0.608         & 0.189 \\
    \cmark & Both & 3 & 
    1.67         & 0.655          & 17.88 & 
    66.38          & 0.524         & 0.157 \\
    \cmark & Both & 6 & 
    \bf{1.25}       & \bf{0.670}      & \bf{18.55} & 
    \bf{67.47}      & \bf{0.423}      & \bf{0.108} \\
    \cmark & Both & 9 & 
    1.59         & 0.659          & 17.95 & 
    66.81         & 0.475          & 0.161 \\
    \bottomrule
  \end{tabularx}
\end{table}

\begin{table}[!t]
  \caption{\textbf{Ablation of Kinematic Estimator.} 
  CD is scaled by $\times 10^{-2}$. 
  Type Acc. (\%) is joint classification accuracy. 
  Axis Err. (rad) and Pivot Err. denote axis direction and pivot distance errors.}
  \label{tab:ablation-ke}
  \vspace{-2 mm}
  \begin{tabularx}{\linewidth}{YYYcYccc}
    \toprule
    \multirow{2}{*}{$\mathbf{Q}_p^L$} &
    \multirow{2}{*}{$\mathbf{H}$} & 
    \multicolumn{3}{c}{Geometry} &
    \multicolumn{3}{c}{Kinematics} \\
    \cmidrule(lr){3-5} \cmidrule(lr){6-8} 
    & & 
    CD \darr & 
    F-Score \uarr & 
    PSNR \uarr & 
    Type Acc. \uarr & 
    Axis Err. \darr & 
    Pivot Err. \darr \\
    \midrule
    \xmark & \cmark & 
    1.65         & 0.652          & 17.92 & 
    67.20          & 0.499          & 0.191 \\
    \cmark & \xmark & 
    2.35       & 0.573      & 18.11 & 
    27.14      & 0.882      & 0.283 \\
    \midrule
    \cmark & \cmark & 
    \bf{1.25}  & \bf{0.670} & \bf{18.55} & 
    \bf{67.47} & \bf{0.423} & \bf{0.108} \\
    \bottomrule
  \end{tabularx}
  \vspace{-4 mm}
\end{table}

\paragraph{Dual-Query Motion Decoder}
As shown in Table~\ref{tab:ablation-qdmd}, we analyze key design choices of the Dual-Query Motion Decoder.
Disabling Dual-Query Initialization (DQI) and randomly initializing $\mathbf{Q}_p^0$ and $\mathbf{Q}_c^0$ degrades both geometry and kinematics, showing the importance of informed query initialization. 
Applying residual updates to only one branch is suboptimal, while updating both position and content queries achieves the best performance, highlighting the need for joint refinement.
Increasing the number of refinement layers improves results up to $L=6$, while a deeper model (\eg, $L=9$) leads to performance degradation.

\paragraph{Kinematic Estimator}
As shown in Table~\ref{tab:ablation-ke}, we ablate the Kinematic Estimator. 
We regress the joint origin using a residual formulation, $\mathbf{m}_o = \mathbf{Q}_p^L + \Delta_o$. 
Directly predicting $\mathbf{m}_o$ without this anchor degrades both geometry and kinematics, showing the benefit of centroid-based residual prediction.
Excluding the point embedding $\mathbf{H}$ also leads to significant performance drops, indicating its importance for parameter regression.
Using both $\mathbf{Q}_p^L$-based residual prediction and $\mathbf{H}$ achieves the best results.

\begin{figure}[!t]
  \includegraphics[width=\textwidth]{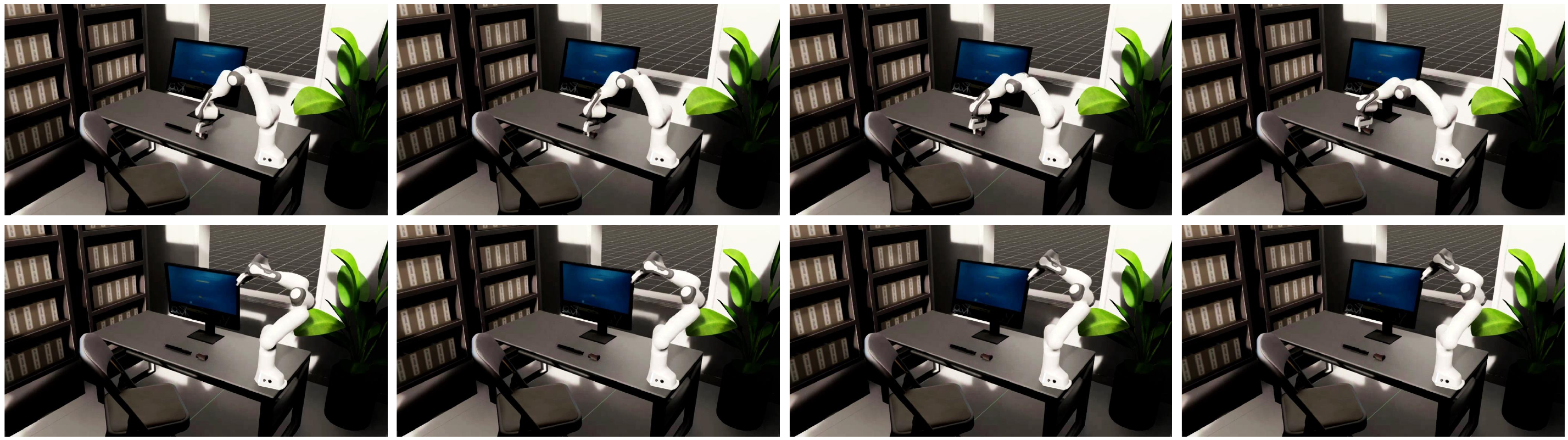}
  \caption{\textbf{Robot manipulation with generated articulated objects.} \OM reconstructions are directly imported into IsaacSim for contact-rich interaction.}
  \label{fig:robotics}
\end{figure}

\begin{figure}[!t]
  \includegraphics[width=\textwidth]{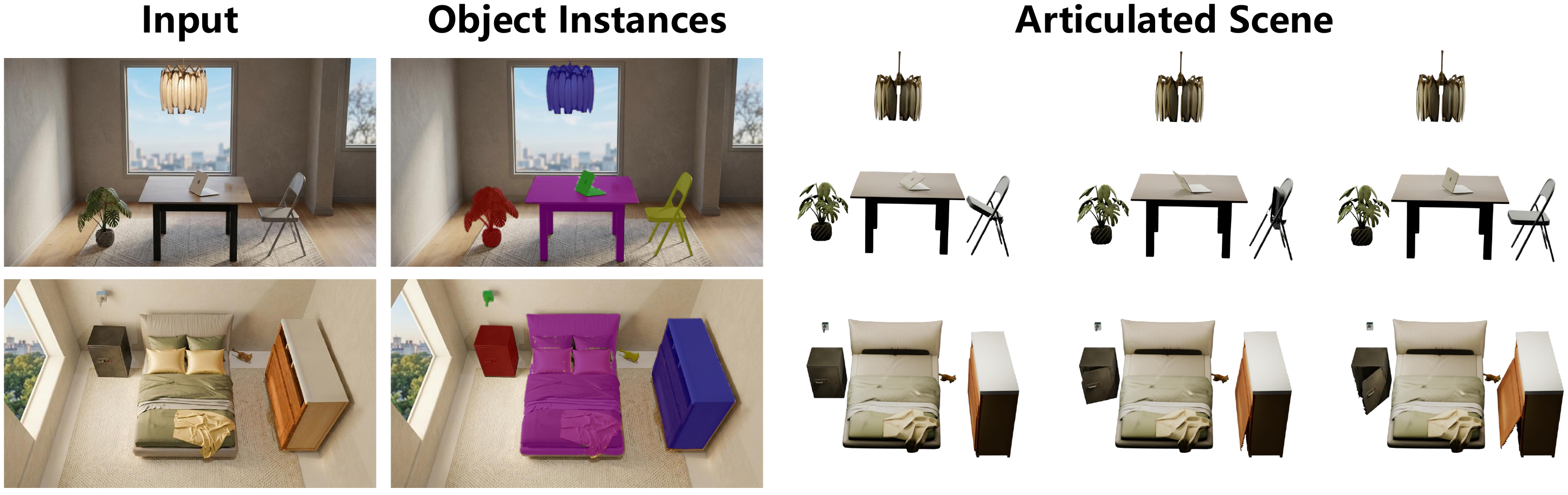}
  \caption{\textbf{Articulated scene reconstruction.} 
  \OM augments SAM3D~\cite{DBLP:journals/corr/abs-2511-16624} with object-level articulation recovery to produce articulated, operable 3D scenes.}
  \label{fig:art-scene}
  \vspace{-4 mm}
\end{figure}

\subsection{Applications}

\paragraph{Robot Manipulation}
\OM extends beyond static 3D reconstruction by inferring articulation axes, joint types, and motion limits that are directly usable for robotic control.
These structured kinematic priors convert monocular observations into actionable parameters, enabling motion reasoning and interaction planning without manual joint annotation.
To evaluate downstream utility, we import the articulated objects reconstructed from in-the-wild images in Fig.~\ref{fig:real-world} into IsaacSim~\cite{DBLP:journals/corr/abs-2511-04831}. 
As shown in Fig.~\ref{fig:robotics}, the simulation-ready assets can be directly manipulated by a Franka robot arm for contact-rich tasks such as grasping and opening, without additional modeling. 
This demonstrates a practical real-to-sim pipeline that produces physically plausible articulated models for robotic interaction and policy learning.

\paragraph{Articulated Scene Reconstruction}
Methods such as MIDI~\cite{DBLP:conf/cvpr/HuangGAY0ZLLCS25} and SAM 3D~\cite{DBLP:journals/corr/abs-2511-16624} provide static scene reconstructions with per-object masks and 6D poses. 
Building on these outputs, we reconstruct each masked object instance with \OM to recover both geometry and articulation parameters. 
The reconstructed articulated objects are then placed back into the scene using their estimated 6D poses, yielding a coherent articulated scene without additional manual modeling.
As shown in Fig.~\ref{fig:art-scene}, this simple object-level augmentation converts rigid scene reconstructions into functionally operable environments, where articulated objects preserve their kinematic structure while remaining consistent with the global layout.

\subsection{Discussion}

\paragraph{Runtime}
All timings are measured on a single NVIDIA A6000 GPU, excluding I/O time, and averaged over 100 runs.
As shown in Fig.~\ref{fig:teaser}, recent methods require 229.9s (Articulate-Anything~\cite{DBLP:conf/iclr/LeXLWYMVKJE25}), 256.8s (PhysXAnything~\cite{DBLP:journals/corr/abs-2511-13648}), 31.6s (PhysXGen~\cite{DBLP:journals/corr/abs-2507-12465}), 34.1s (URDFormer~\cite{DBLP:conf/rss/ChenWMMFF024}), and 19.6s (SINGAPO~\cite{DBLP:conf/iclr/LiuICSA25}) per instance.
In comparison, \OM requires 20.5 seconds per instance. 
Among this, 18.2s is spent on TRELLIS-based 3D reconstruction, while articulation reasoning and post-processing introduce only marginal overhead.

\paragraph{Limitations}
While \OM demonstrates strong performance in articulated object reconstruction and motion reasoning, several limitations remain.
%
\OM can struggle with very small parts attached to large objects (\eg, tiny buttons). 
Due to uniform point sampling over the entire shape, such components may receive only sparse coverage, making their features less distinctive and more prone to over-smoothing. 
Consequently, articulations under extreme scale imbalance can be difficult to reliably segment and parameterize.
%
\OM also relies on learned structural priors over part–whole relationships, which may not fully generalize to objects with novel topologies or uncommon articulation patterns. 
For such unseen object configurations, the predicted motion parameters (\OM, axes or ranges) can be less accurate, even when part segmentation remains reasonable.

\section{Conclusion}

In this work, we present \textbf{\OM}, a unified framework for monocular articulated 3D reconstruction grounded in progressive structural reasoning.
Instead of depending on multi-view supervision, retrieval libraries, or auxiliary video synthesis, \OM formulates monocular articulated reconstruction as a progressive structural reasoning process.
By explicitly modeling geometry, part structure, and motion in a unified framework, it achieves accurate and efficient articulation inference without handcrafted motion priors or external pipelines.
Extensive experiments demonstrate that \OM achieves state-of-the-art performance in both reconstruction accuracy and inference speed on PartNet-Mobility.
Beyond object-level reconstruction, the framework generalizes effectively to robotic manipulation and articulated scene reconstruction, highlighting its practical applicability. 

\ifreview \else
\section*{Acknowledgements}
This study is supported by the Ministry of Education, Singapore, under its MOE AcRF Tier 2 (MOET2EP20221-0012, MOE-T2EP20223-0002), and by cash and in-kind contributions from NTU S-Lab and industry partner(s).
\fi

\bibliographystyle{splncs04}
\bibliography{references}

\appendix
\onecolumn
\section{Implementation Details}

\subsection{Loss Functions}

We supervise the model with five training objectives: a triplet loss $\ell_{\text{triplet}}$ for the Part-Aware Semantic Reasoner, a mask loss $\ell_{\text{mask}}$ and a confidence loss $\ell_{\text{score}}$ for Dual-Query Motion Decoder, a motion loss $\ell_{\text{motion}}$ for articulation parameter regression, and a structure loss $\ell_{\text{struct}}$ for kinematic tree prediction.

To supervise the fixed set of predicted queries, we establish a one-to-one correspondence between predicted queries and ground-truth articulated parts using Hungarian bipartite matching. The matching cost is computed from mask similarity between the predicted part mask and the ground-truth part mask, using a weighted combination of binary cross-entropy and Dice cost. The resulting assignment identifies the queries matched to ground-truth articulated parts, while the remaining queries are treated as null predictions. 

\paragraph{Part-Aware Semantic Reasoner}
The Part-Aware Semantic Reasoner is supervised by a triplet contrastive loss to learn discriminative motion-aware part embeddings. 
For a triplet $(\mathbf{h}_a,\mathbf{h}_b,\mathbf{h}_c)$, where $(\mathbf{h}_a,\mathbf{h}_b)$ belong to the same articulated part and $\mathbf{h}_c$ belongs to a different part, the triplet loss is defined as
\begin{equation}
\ell_{\text{triplet}}
=
-\frac{1}{2}
\Bigl[
\log\frac{s_{ab}}{s_{ab}+s_{ac}}
+
\log\frac{s_{ab}}{s_{ab}+s_{bc}}
\Bigr],
\end{equation}
where the pairwise similarity score is defined as
\begin{equation}
s_{ij}
=
\exp\left(
\frac{\cos(\mathbf{h}_i,\mathbf{h}_j)}{\tau}
\right),
\end{equation}
and $\tau$ is a learnable temperature parameter.

\paragraph{Dual-Query Motion Decoder}
The Dual-Query Motion Decoder is supervised by three terms: a \textbf{mask loss} for part segmentation, a \textbf{confidence loss} for query reliability estimation, and an \textbf{auxiliary object-category classification loss} for query initialization. 

\noindent \textbf{Mask loss} is defined for each query matched to a ground-truth part, where the predicted mask $m_q$ is supervised against its matched ground-truth mask $m_q^{gt}$ using focal and Dice losses:
\begin{equation}
\ell_{\text{mask}}
=
\lambda_{\text{focal}}\ell_{\text{focal}}
+
\lambda_{\text{dice}}\ell_{\text{dice}}.
\end{equation}

\noindent \textbf{Confidence loss} is applied to the predicted confidence score $\hat{c}_q$, which indicates the reliability of the predicted part mask. The supervision target is defined as the mask IoU between the predicted mask and the matched ground-truth mask:
\begin{equation}
u_q
=
\mathrm{IoU}(m_q,m_q^\text{gt}).
\end{equation}
Following the Quality Focal Loss formulation, the confidence loss is
\begin{equation}
\ell_{\text{score}}
=
\left|
\sigma(\hat{c}_q)-u_q
\right|^\beta
\cdot
\mathrm{BCE}(\hat{c}_q,u_q),
\end{equation}
where $\sigma(\cdot)$ denotes the sigmoid function. Queries not matched to any ground-truth part are assigned zero targets. 
During inference, queries with predicted confidence scores below $0.5$ are filtered out, and the remaining queries are retained as valid articulated part hypotheses. 

\noindent \textbf{Auxiliary object-category classification} is introduced to supervise the dual-query initialization branch:
\begin{equation}
\ell_{\text{obj}}
=
\mathrm{CE}(\hat{\mathbf{y}}_{\text{obj}},\mathbf{y}_{\text{obj}}^{gt}).
\end{equation}
This auxiliary supervision stabilizes dual-query initialization in early training.

\paragraph{Articulation Parameter Regressor}
The Articulation Parameter Regressor is supervised by a motion loss that includes joint type, axis direction, motion origin, and motion limit regression:
\begin{equation}
\ell_{\text{motion}}
=
\lambda_t\ell_{\text{type}}
+
\lambda_d\ell_{\text{dir}}
+
\lambda_o\ell_{\text{origin}}
+
\lambda_l\ell_{\text{limit}}.
\end{equation}
where, the joint type loss is
\begin{equation}
\ell_{\text{type}}
=
\mathrm{CE}(\hat{\mathbf{t}}_q,\mathbf{t}_q^{gt}).
\end{equation}
The axis direction loss is defined using unsigned cosine similarity:
\begin{equation}
\ell_{\text{dir}}
=
1-
\left|
\frac{\hat{\mathbf{a}}_q}{\|\hat{\mathbf{a}}_q\|}
\cdot
\frac{\mathbf{a}_q^{gt}}{\|\mathbf{a}_q^{gt}\|}
\right|.
\end{equation}
The motion origin loss is
\begin{equation}
\ell_{\text{origin}}
=
\|
\hat{\mathbf{o}}_q - \mathbf{o}_q^{gt}
\|_1.
\end{equation}
For joints with bounded motion ranges, we use a center--span parameterization. Let $l_{\min}$ and $l_{\max}$ denote the lower and upper motion bounds, and define
\begin{equation}
c = \frac{l_{\min} + l_{\max}}{2}, \qquad
s = \frac{l_{\max} - l_{\min}}{2}.
\end{equation}
The network predicts $\hat{c}_q$ and $\hat{s}_q$, and the corresponding loss is
\begin{equation}
\ell_{\text{limit}}
=
\|
\hat{c}_q - c_q^{gt}
\|_1
+
\|
\hat{s}_q - s_q^{gt}
\|_1.
\end{equation}

\paragraph{Kinematic Tree Predictor}
The Kinematic Tree Predictor is supervised by a structure loss for parent prediction, computed over queries matched to ground-truth articulated parts:
\begin{equation}
\ell_{\text{struct}}
=
\frac{1}{N_{\text{match}}}
\sum_{q \text{ matched}}
\mathrm{CE}(\hat{\mathbf{s}}_q,p_q^{gt}),
\end{equation}
where $N_{\text{match}}$ denotes the number of queries matched to ground-truth articulated parts, $\hat{\mathbf{s}}_q$ denotes the predicted parent probability distribution, and $p_q^{gt}$ is the ground-truth parent index.

\subsection{Training Details}

\paragraph{Training Procedure}
Training is performed on four NVIDIA A100 GPUs with a batch size of 1 per GPU (effective batch size of 4) and requires approximately six days. 
The training procedure consists of four stages.

\noindent \textbf{Stage I} warms up the Part-Aware Semantic Reasoner for 100 epochs using only the triplet loss while freezing the TRELLIS backbone and all downstream modules:
\begin{equation}
\ell_{\text{stage1}} = \ell_{\text{triplet}}.
\end{equation}

\noindent \textbf{Stage II} freezes the semantic reasoner and trains the dual-query initialization branch for 20 epochs with object-category supervision:
\begin{equation}
\ell_{\text{stage2}} = \ell_{\text{obj}}.
\end{equation}

\noindent \textbf{Stage III} jointly optimizes the Part-Aware Semantic Reasoner, Dual-Query Motion Decoder, and Articulation Parameter Regressor for 100 epochs using
\begin{equation}
\ell_{\text{stage3}}
=
\lambda_{\text{triplet}} \ell_{\text{triplet}}
+
\lambda_{\text{mask}} \ell_{\text{mask}}
+
\lambda_{\text{score}} \ell_{\text{score}}
+
\lambda_{\text{motion}} \ell_{\text{motion}}.
\end{equation}
The weight of $\ell_{\text{motion}}$ is linearly increased during the first 40 epochs. 

\noindent \textbf{Stage IV} freezes all preceding modules and trains the Kinematic Tree Predictor for 30 epochs using
\begin{equation}
\ell_{\text{stage4}} = \ell_{\text{struct}}.
\end{equation}

\begin{figure}[!t]
    \includegraphics[width=\linewidth]{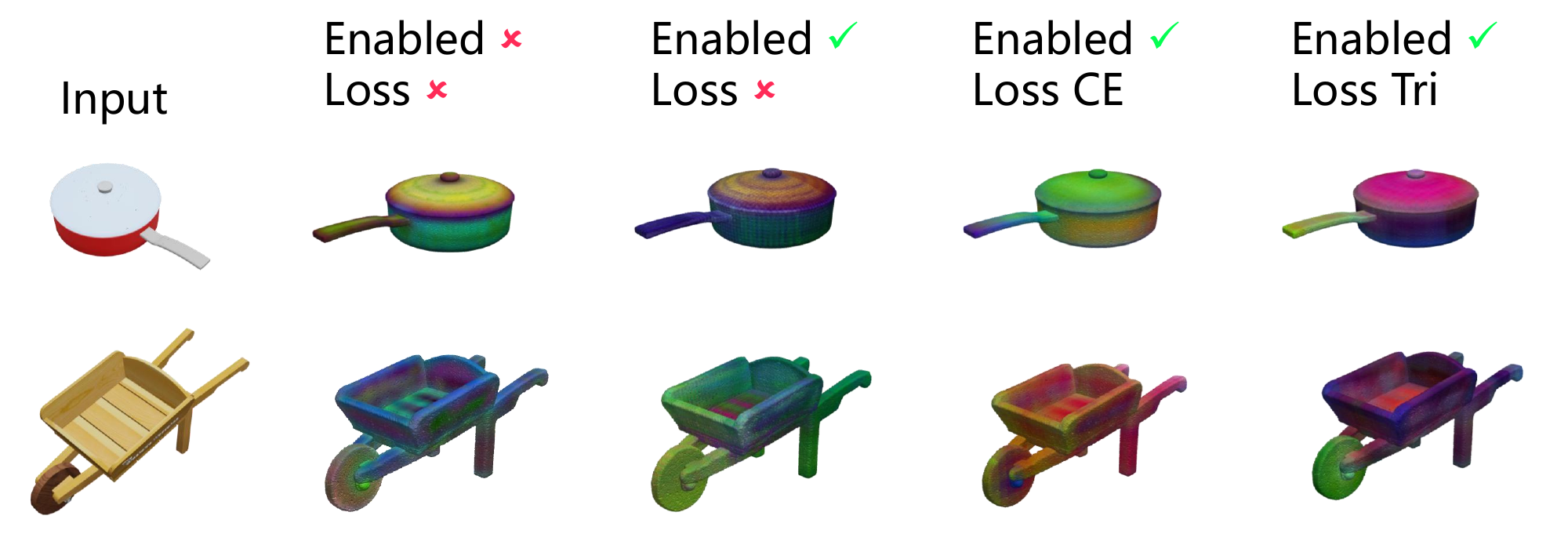}
    \caption{
        \textbf{Visualization of point features $\mathbf{H}$ from the Part-Aware Semantic Reasoner.}
        Points are colored by PCA projection of the features.
        ``Enabled'' indicates whether the reasoner is used, and ``Loss'' specifies the supervision type.
        Note that ``CE'' and ``Tri.'' denote ``Cross-Entropy'' and ``Triplet'', respectively.
        }
    \label{fig:pasr}
\end{figure}

\begin{figure}[!t]
    \centering
    \includegraphics[width=\linewidth]{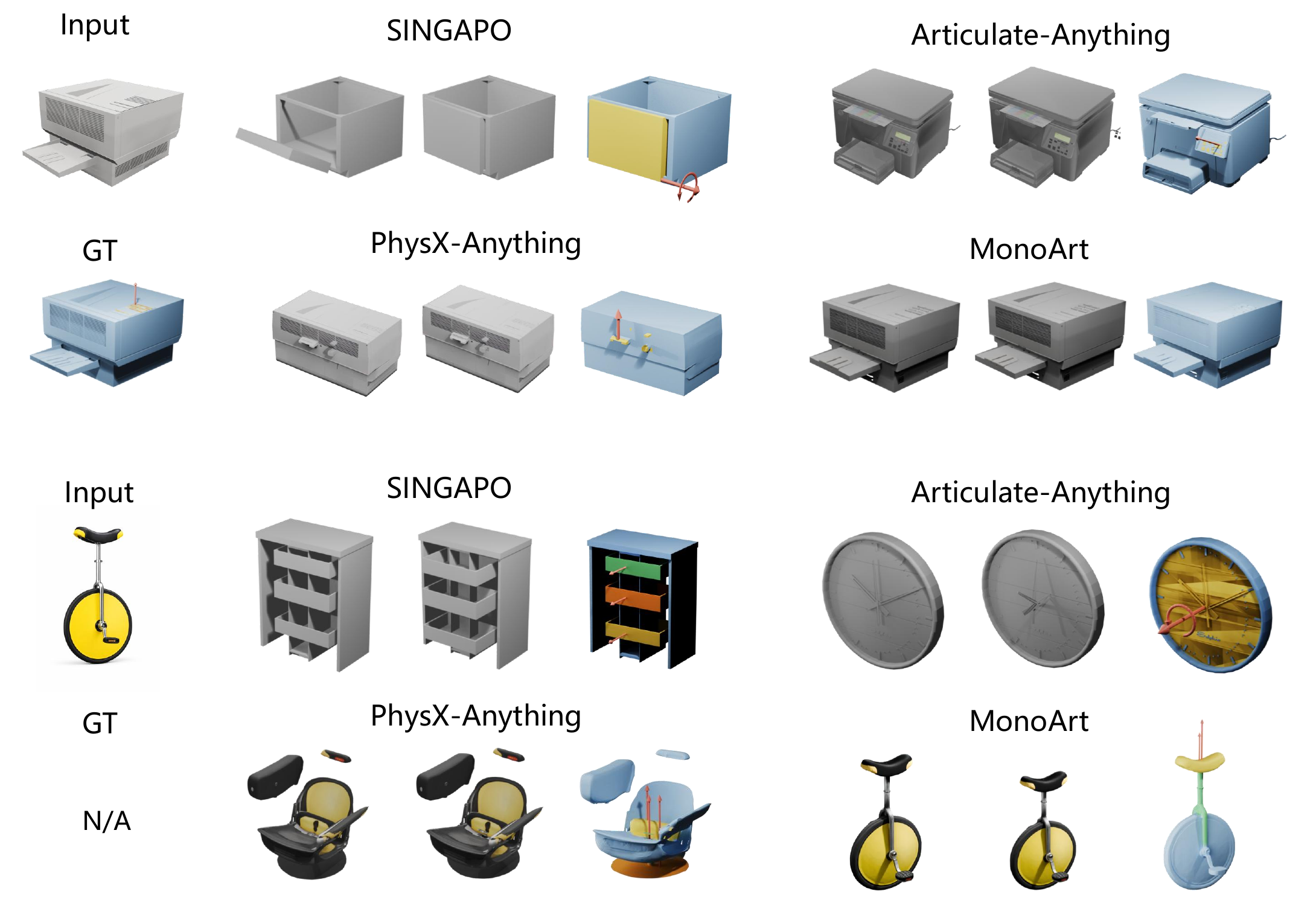}
    \caption{\textbf{Representative Failure Cases.} (Top) an object containing an extremely small component (printer button). (Bottom) an unseen articulated object (bicycle).}
    \label{fig:failure-cases}
\end{figure}

\paragraph{Loss Weights}
Unless otherwise specified, we use $\lambda_{\text{triplet}}=0.2$, $\lambda_{\text{mask}}=1.0$, $\lambda_{\text{score}}=1.0$, and $\lambda_{\text{motion}}=1.0$. For mask supervision, we use $\lambda_{\text{focal}}=1.0$ and $\lambda_{\text{dice}}=1.0$ with focal parameter $\gamma=2$. For confidence score supervision, we use $\beta=2$. For motion regression, we use $\lambda_t=1.0$, $\lambda_d=1.0$, $\lambda_o=1.0$, and $\lambda_l=1.0$. For Hungarian matching, we use equal weights for BCE and Dice costs.

\paragraph{Optimizer}
All stages use the AdamW optimizer with weight decay $0.01$. Mixed-precision training (FP16) is used to reduce memory usage and accelerate training. Gradients are clipped with a maximum norm of $0.1$.

\paragraph{Learning Rate Schedule}
The learning rate is warmed up for 10 epochs from $1\%$ of the base rate ($5\times10^{-5}$), followed by cosine annealing to $10^{-6}$.

\section{Additional Ablation Study Results}

\subsection{Part-Aware Semantic Reasoner}

Fig.~\ref{fig:pasr} qualitatively complements the ablation results in Table~\ref{tab:ablation-pasr} by visualizing the learned point features under different variants of Part-Aware Semantic Reasoner (PASR).
Without PASR, the features show limited part discrimination. 
Adding PASR improves structural organization, while different supervision strategies further affect feature separability. 
Cross-entropy supervision yields partially separated clusters, whereas triplet supervision produces compact and well-separated part features that better align with articulated components.

\section{More Discussion on Limitations}

Fig.~\ref{fig:failure-cases} presents two representative failure cases, revealing key challenges in monocular articulated reconstruction.
\textbf{1) Extremely small components} (\eg, the printer button) are difficult to capture under uniform sampling and limited spatial resolution, resulting in inaccurate part identification and placement.
\textbf{2) Unseen articulated categories} (\eg, bicycle) lead to incorrect articulation recovery due to the large domain gap from the training data.

\section{More Qualitative Results}

\begin{figure}[t]
    \includegraphics[width=\linewidth]{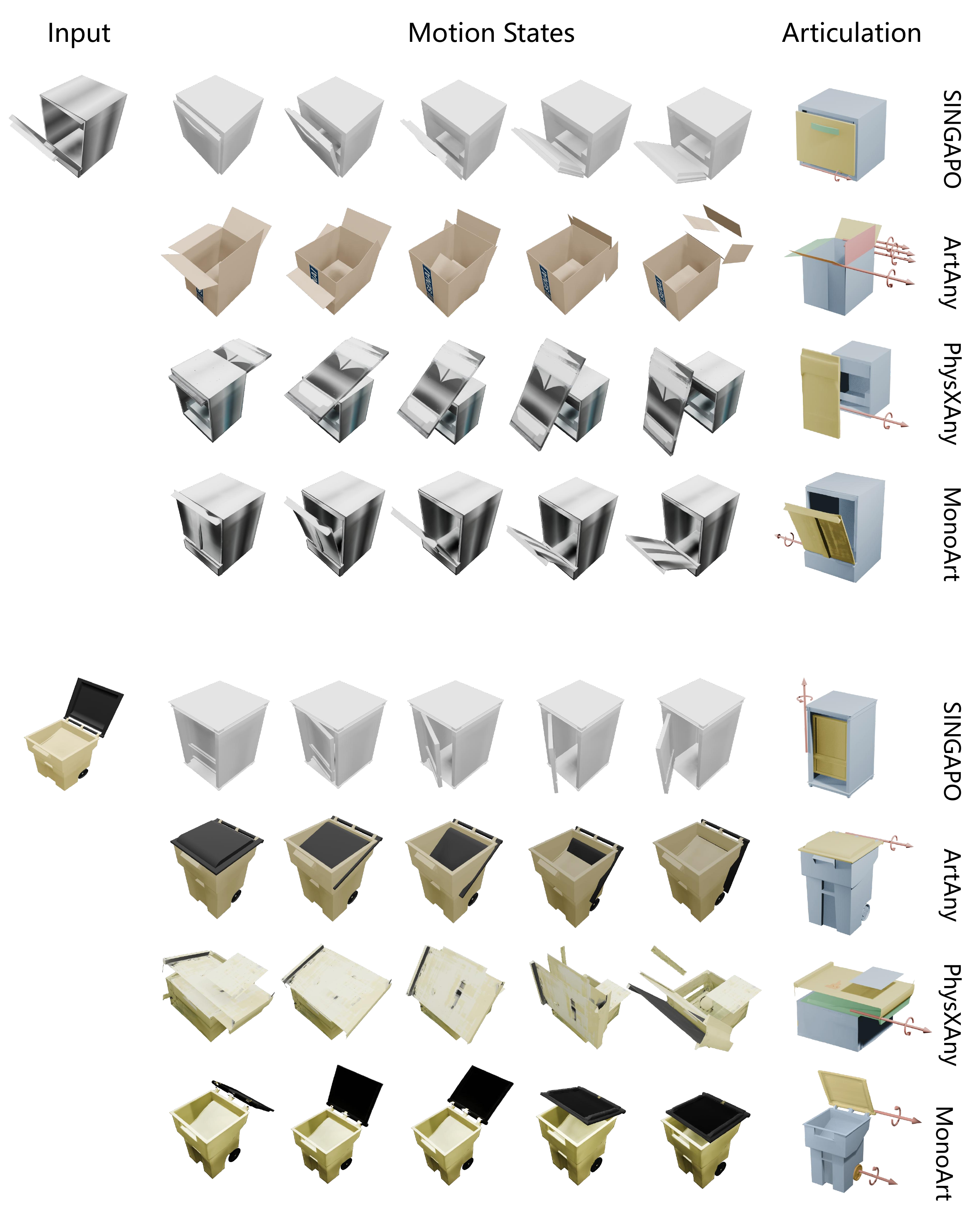}
    \caption{\textbf{Additional qualitative results on articulated object reconstruction.}
    We compare SINGAPO~\cite{DBLP:conf/iclr/LiuICSA25}, Articulate-Anything (ArtAny)~\cite{DBLP:conf/iclr/LeXLWYMVKJE25}, PhysX-Anything (PhysXAny)~\cite{DBLP:journals/corr/abs-2511-13648}, and \OM across multiple motion states and visualize the predicted articulations (part masks and motion parameters).}
    \label{fig:More_results1}
\end{figure}

\begin{figure}[t]
    \includegraphics[width=\linewidth]{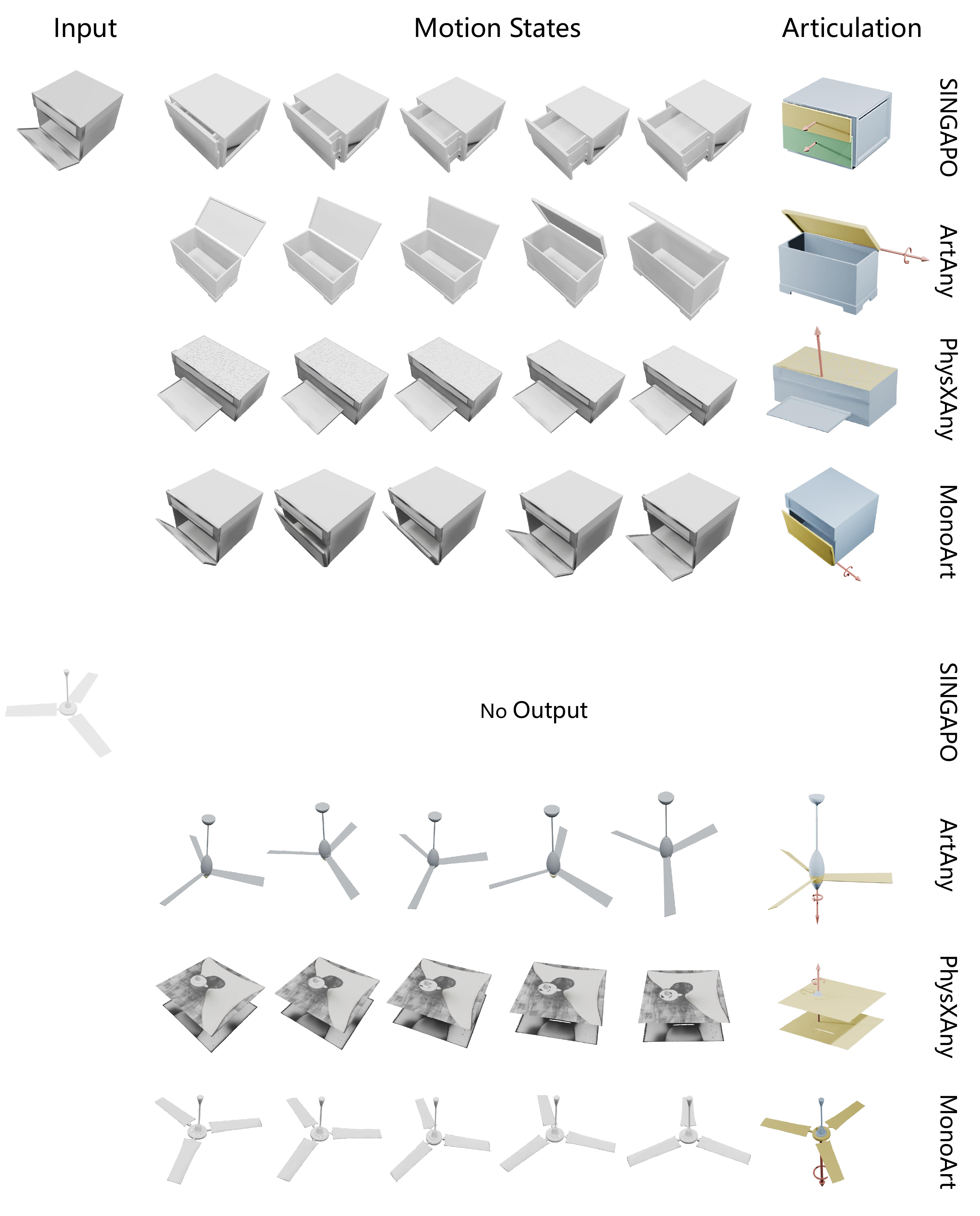}
    \caption{\textbf{Additional qualitative results on articulated object reconstruction.}
    We compare SINGAPO~\cite{DBLP:conf/iclr/LiuICSA25}, Articulate-Anything (ArtAny)~\cite{DBLP:conf/iclr/LeXLWYMVKJE25}, PhysX-Anything (PhysXAny)~\cite{DBLP:journals/corr/abs-2511-13648}, and \OM across multiple motion states and visualize the predicted articulations (part masks and motion parameters).}
    \label{fig:More_results2}
\end{figure}

We present additional qualitative comparisons on articulated object reconstruction in Fig.~\ref{fig:More_results1} and ~\ref{fig:More_results2}. 
Across diverse objects and motion ranges, \OM reconstructs more consistent geometry and more plausible articulated motion than prior methods.

\end{document}